\documentclass{article}
\usepackage{ijcai22}
\usepackage{multicol}
\usepackage{multirow}
\usepackage{amsfonts,amssymb}
\usepackage{times}
\usepackage{soul}
\usepackage{url}
\usepackage[hidelinks]{hyperref}
\usepackage[utf8]{inputenc}
\usepackage[small]{caption}
\usepackage{graphicx}
\usepackage{amsmath}
\usepackage{amsthm}
\usepackage{booktabs}
\usepackage{algorithm}
\usepackage{algorithmic}
\title{FastRE: Towards Fast Relation Extraction with Convolutional Encoder and Improved Cascade Binary Tagging Framework}

\author{
Guozheng Li$^1$
\and
Xu Chen$^4$\and
Peng Wang${^1}{^,}{^2}{^,}{^3}$\thanks{Corresponding author} \and
Jiafeng Xie$^1$ \And
Qiqing Luo${^1}{^,}{^3}$
\affiliations
$^1$School of Computer Science and Engineering, Southeast University\\
$^2$School of Cyber Science and Engineering, Southeast University\\
$^3$School of Artificial Intelligence, Southeast University\\
$^4$Tencent Inc.\\
\emails
\{gzli, pwang, xjf, qiqingluo\}@seu.edu.cn, bigxuchen@tencent.com
}

\begin{document}

\maketitle

\begin{abstract}

Recent work for extracting relations from texts has achieved excellent performance. 
However, most existing methods pay less attention to the efficiency, making it still challenging to quickly extract relations from massive or streaming text data in realistic scenarios. 
The main efficiency bottleneck is that these methods use a Transformer-based pre-trained language model for encoding, which heavily affects the training speed and inference speed. 
To address this issue, we propose a fast relation extraction model (FastRE) based on convolutional encoder and improved cascade binary tagging framework. 
Compared to previous work, FastRE employs several innovations to improve efficiency while also keeping promising performance. 
Concretely, FastRE adopts a novel convolutional encoder architecture combined with dilated convolution, gated unit and residual connection, which significantly reduces the computation cost of training and inference, while maintaining the satisfactory performance. 
Moreover, to improve the cascade binary tagging framework, FastRE first introduces a type-relation mapping mechanism to accelerate tagging efficiency and alleviate relation redundancy, and then utilizes a position-dependent adaptive thresholding strategy to obtain higher tagging accuracy and better model generalization. 
Experimental results demonstrate that FastRE is well balanced between efficiency and performance, and achieves 3-10$\times$ training speed, 7-15$\times$ inference speed faster, and 1/100 parameters compared to the state-of-the-art models, while the performance is still competitive. Our code is available at \url{https://github.com/seukgcode/FastRE}.


\end{abstract}

\section{Introduction}
Relation extraction (RE) aims to identify the relationships between entities in texts, and plays an important role in natural language processing applications, knowledge graph construction, and question answering.
Recent RE studies have made great progress in pursuing excellent performance~\cite{wei2020novel,ren-etal-2021-novel}. 
However, in realistic scenarios, RE models are often expected to keep high performance and fast speed simultaneously.
For example, financial investors require extracting relational triples effectively from massive and streaming real-time news, financial and political data to build knowledge graphs~\cite{dong2014knowledge} to help make decisions. 
Unfortunately, most existing methods lack considering time consumed in model training and inference and cannot efficiently extract relations from texts. 
Early RE research applied approximate frequency counting and dimension reduction to speed up similarity computation in unsupervised RE~\cite{takase2015fast}, and adopted neural metric learning methods to accelerate RE~\cite{tran2019neural}.
Various CNN based approaches~\cite{zeng2014relation,santos2015classifying} had also shown effective in tackling this problem.
However, these efforts fail to address the challenges of RE on both performance and efficiency.

Recent advances on RE performance~\cite{wei2020novel,wang2020tplinker,sui2021joint,zheng-etal-2021-prgc,ren-etal-2021-novel,huguet-cabot-navigli-2021-rebel-relation} mainly due to adopting the encoders with the Transformer~\cite{vaswani2017attention} based pre-trained language models (PLMs) such as BERT~\cite{devlin2018bert} and BART~\cite{lewis-etal-2020-bart}, which have powerful abilities to capture long-distance dependencies and context semantic features. 
However, the token-pair based attention operation in Transformer requires much time consumption and memory consumption within GPUs. 
Moreover, the memory consumption of PLMs restricts the batch size during model training and inference, which means that models are restricted to set a relatively small batch size within limited computing resources. 
Although it is not a serious problem in training, it limits the parallel processing ability in the inference.


Instead of using the Transformer, we design a novel convolutional structure to tackle the computational efficiency issue in encoder. 
It significantly accelerates the training and inference speed with dilated convolution~\cite{yu2015multi}, gated unit~\cite{dauphin2017language} and residual connection~\cite{he2016deep}. 
First, the dilated convolution increases the receptive region of network outputs exponentially with respect to the network depth, which results in obtaining drastically shortened computation paths and capturing arbitrarily long-distance dependencies.
In other words, using dilated convolution can achieve the high efficiency of vanilla convolution with fewer layers. 
Second, gated unit is used to control what information should be propagated through the hierarchy of layers. 
Then residual connection is used for avoiding gradient vanishing to enable deep convolutional networks. 
Our convolutional encoder not only greatly reduces the model training and inference consumed time, but also ensures the competitive performance in RE.

On the other side, recent work also shows the effectiveness of the cascade binary tagging framework~\cite{wei2020novel} for solving the overlapping RE. 
However, this framework suffers from two shortcomings: relation redundancy and poor generalization~\cite{zheng-etal-2021-prgc}. 
Relation redundancy, namely, extracting tail entity extraction to all relations, would cause plenty of meaningless computations. 
Besides, due to that the cascade binary tagging framework reduces the multi-label problem to a binary classification problem, 
it needs heuristic threshold tuning and introduces boundary decision errors. 

To solve above issues, we introduce an entity type to predefined relation (type-relation) mapping mechanism and a position-dependent adaptive thresholding strategy to improve the cascade binary tagging framework. 
Specifically, the mapping between head entity type and predefined relation is maintained, when the head entity type is determined, its potential corresponding relations are also determined. 
Under this mechanism, it avoids traversing for all the relations when predicting overlapping relations. 
In addition, incorporating entity type information into RE can improve performance~\cite{zhong-chen-2021-frustratingly}. 
The position-dependent adaptive thresholding replaces the global threshold with learnable thresholds for different positions in a sentence when performing binary tagging. 
The thresholds are learned with a rank-based loss function~\cite{zhou2021document}, that pushes positive classes scores above the thresholds and pulls negative classes scores below in training. 
And the tagger sets a position to 1 that has the higher score than its position-dependent threshold or set to 0 that has the lower score. 
This simple strategy avoids threshold tuning, and makes the threshold adjustable to different tagging positions, which leads to better generalization.

In this paper, we propose a simple and fast relation extraction model, FastRE, to significantly reduce training and inference time but retains competitive performance. 
To the best of our knowledge, FastRE is the first effort to address the balance between efficiency and performance in RE. 
In summary, the main contributions of this paper are three-fold: 
\begin{itemize}
\item We propose a novel convolutional encoder combined with dilated convolution, gated unit and residual connection for RE, which significantly reduces the computation cost but maintaining the satisfactory performance. 
\item We improve the cascade binary tagging framework with type-relation mapping mechanism and position-dependent adaptive thresholding, which solves the relation redundancy and poor generalization issues.
\item Experimental results on public datasets show that FastRE achieves 3-10$\times$ training speed, 7-15$\times$ inference speed faster, and 1/100 parameters compared to the state-of-the-art models, while the performance still competitive, namely, it is even slightly improved on NYT10 and NYT11, and only reduced by less than 4\% on NYT24. 
\end{itemize}


\section{Methodology}
\subsection{Problem Formulation and Model Overview}
Given the predefined relation set $\mathcal R = \{r_1, r_2,..., r_N\}$, for any sentence $\mathcal S$ containing the entity set $\mathcal E = \{e_1, e_2,..., e_M\}$ with the entity type set $\mathcal T = \{t_1, t_2,..., t_K\}$, the relation extraction task aims to extract the relational triples in $\mathcal Z = \{(e_i, r_k, e_j)\}$, where $N$ denotes the number of relations, $M$ represents the number of entities, $K$ is the number of entity types, and $e_i, e_j \in \mathcal E, r_k \in \mathcal R$. 

\begin{figure*}[ht]
	\centering
	\includegraphics[width=0.9\textwidth]{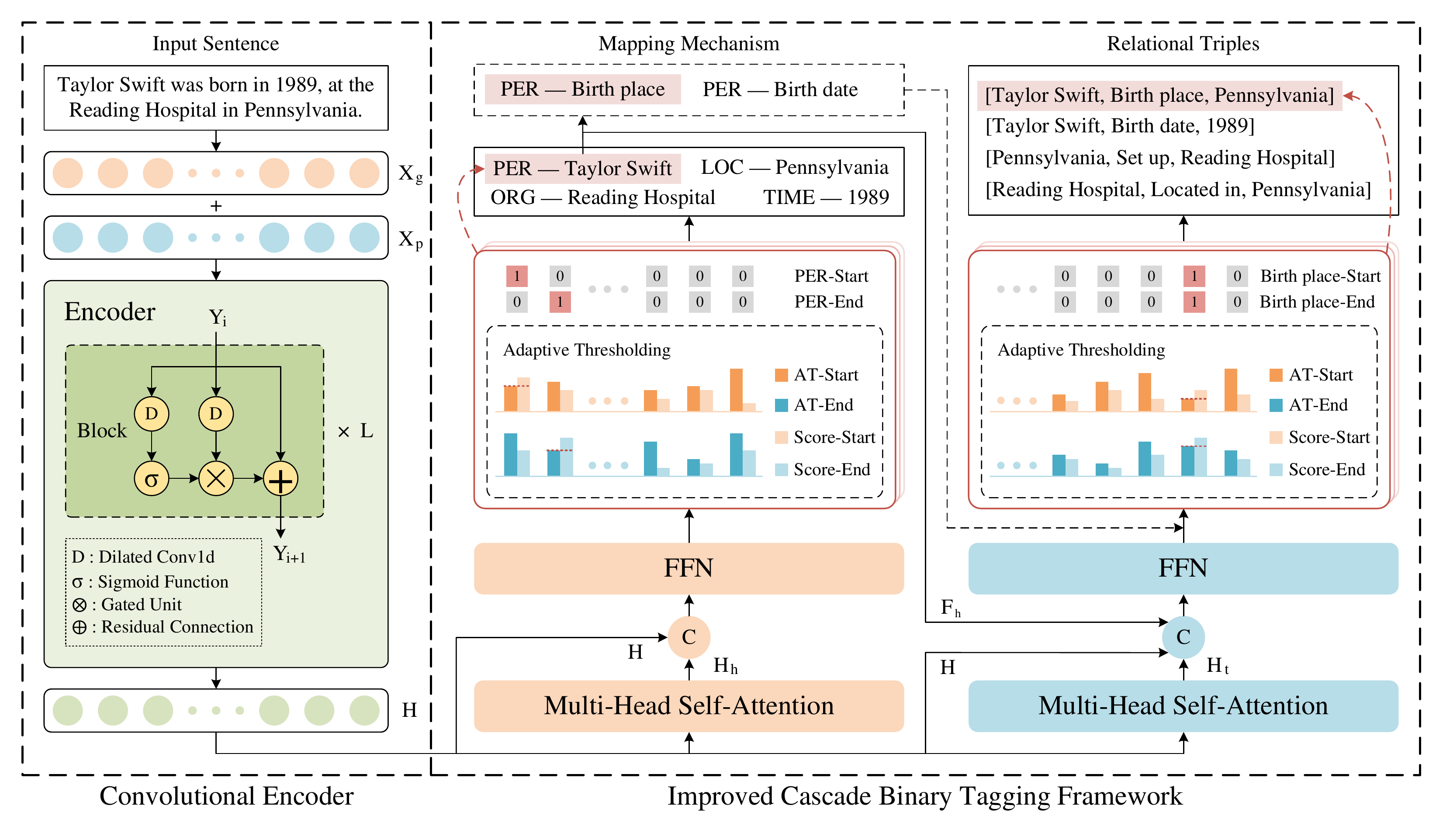}
	\caption{The overall structure of FastRE. 
	}
	\label{overview}
\end{figure*}

The principle of FastRE is shown in Figure~\ref{overview}. 
FastRE consists of a convolutional encoder and a improved cascade binary tagging framework. 
The convolutional encoder contains $L$ stacked $\rm{Block}(\cdot)$ which consists of two dilated convolutions: a gated unit and a residual connection. 
Compared with the original one, the improved cascade binary tagging framework not only adds type-relation mapping mechanism to determine potential relations based on entity types, but also adopts position-dependent adaptive thresholding strategy for different positions.
First, the convolutional encoder efficiently converts the input sentence $\mathbf{X}$ composed of word embedding $\mathbf{X_g}$ and position embedding $\mathbf{X_p}$ into sentence representation $\mathbf{H}$. 
Subsequently, two separate auxiliary features $\mathbf{H}_h$ and $\mathbf{H}_t$ are generated based on $\mathbf{H}$ via two different multi-head self-attention layers.
Then FastRE utilizes a feed forward network (FFN) to obtain all the head entities and their types. 
Concretely, sentence representation $\mathbf{H}$ and head entity auxiliary features $\mathbf{H}_h$ are concatenated and feed to the FNN. Then, for each entity type, FastRE calculates the score of every token as the start and end position of a head entity. With adaptive thresholding, the position that has higher score than its position-dependent threshold is set to 1.
Through the mapping mechanism, FastRE determines the potential relations corresponding to the current head entity type. 
Finally, similar to head entity tagging, with the sentence representation $\mathbf{H}$, tail entity auxiliary features $\mathbf{H}_t$ and head entity features $\mathbf{F}_h$, FastRE utilizes another FFN to obtain all the tail entities and forms all the relational triples in $\mathcal Z$.

\subsection{Convolutional Encoder}
Theoretically, we can achieve the aim of modeling arbitrarily long-distance dependencies similar to Transformer~\cite{vaswani2017attention} by stacking enough vanilla convolutional layers. But too many parameters will bring too much computation.
Therefore, we choose the dilated convolution~\cite{yu2015multi} instead because increasing the receptive region in convolution can achieve the high efficiency of vanilla convolution with fewer layers. However, the pre-determined receptive region prevents subsequent layers from examining previous information in detail. We alleviate this issue by using the gated unit~\cite{dauphin2017language} in convolutions to select important features of low-level tokens and phrases. To enable deep networks, we adopt the residual connection~\cite{he2016deep} to avoid gradient vanishing.

Let $\mathbf{X} = [\mathbf{x}_1, \mathbf{x}_2,..., \mathbf{x}_n]$ represent the input sentence, where $\mathbf{x}_i \in \mathbb{R}^d$ is the $i$-th token embedding with dimension $d$, composed of GloVe embedding~\cite{pennington2014glove} $\mathbf{X_g}$ and trainable position embedding $\mathbf{X_p}$. 
The encoder contains $L$ stacked $\rm{Block}(\cdot)$. 
The sentence representation $\mathbf{H} = [\mathbf{w}_1, \mathbf{w}_2,..., \mathbf{w}_n]$ is obtained by:
\begin{equation}
    \mathbf{H} = \rm{Block}(\cdot \cdot \cdot(\rm{Block}(\mathbf{X})))
\label{sentencerepresentation}
\end{equation}
where $\mathbf{H} \in \mathbb{R}^{n \times d}$ and $\mathbf{w}_i \in \mathbb{R}^d$ is the encoded contextual representation of $i$-th token. The $i$-th $\rm{Block}(\cdot)$ in the encoder contains two dilated convolutions with dilation rate $d_i$, a gated unit, and a residual connection. It contains $d$ convolution kernels with size $k_s$. We denote dilated convolution as $\rm{DilatedConv}(\cdot)$ to map the input $\mathbf{X} \in \mathbb{R}^{n \times d}$ to the output $\mathbf{Y}_{a, b} \in \mathbb{R}^{n \times d}$:
\begin{eqnarray}
    \mathbf{Y}_a &=& {\rm{DilatedConv}}_{a}(\mathbf{X}) \\
    \mathbf{Y}_b &=& {\rm{DilatedConv}}_{b}(\mathbf{X})
\end{eqnarray}
For all the $\rm{Block}(\cdot)$, we ensure the output dimension matches the input dimension by padding. 
Similar to~\cite{gehring2017convolutional}, we implement a gated unit over $\mathbf{Y}_a, \mathbf{Y}_b$ and add a residual connection from the $\rm{Block}(\cdot)$ input to its output:
\begin{equation}
    \mathbf{Y}_i = \mathbf{Y}_a \otimes {\rm{sigmoid}}(\mathbf{Y}_b) + \mathbf{X}
    \label{4}
\end{equation}
where $\otimes$ denotes the element-wise multiplication, $\mathbf{Y}_i$ is the output of $i$-th block and the input of $(i+1)$-th block. Obviously, the sentence representation $\mathbf{H}$ is equal to $\mathbf{Y}_L$, which is the output of the last $\rm{Block}(\cdot)$.

\subsection{Improved Cascade Binary Tagger}
FastRE first tags all the head entities, then tags their corresponding relations and tail entities, which is a typical cascade binary tagging framework~\cite{wei2020novel}. 
Furthermore,  FastRE introduces the type-relation mapping mechanism and position-dependent adaptive thresholding to overcome the drawbacks (i.e., relation redundancy and poor generalization~\cite{zheng-etal-2021-prgc}) of cascade binary tagging framework.

Because the convolutional encoder shares the most parameters, it treats every token in the sentence equally and pays less attention to the most important parts of the sentence. Therefore, FastRE generates two separate auxiliary features via two multi-head self-attention~\cite{vaswani2017attention} layers for head entity tagging and tail entity tagging, respectively. Take the head entity auxiliary features $\mathbf{H}_h$ as the example:
\begin{eqnarray}
    \mathbf{H}_h &=& {\rm{softmax}}(\frac{\mathbf{Q}\mathbf{K^ \top}}{\sqrt{d_k}})\mathbf{V} \\
    \mathbf{Q} &=& \mathbf{W}_q \cdot \mathbf{H} + \mathbf{b}_q \\
    \mathbf{K} &=& \mathbf{W}_k \cdot \mathbf{H} + \mathbf{b}_k \\
    \mathbf{V} &=& \mathbf{W}_v \cdot \mathbf{H} + \mathbf{b}_v
\end{eqnarray}
where $d_k = d$ is the dimension of the attention key. The $\mathbf{W}_q, \mathbf{W}_k, \mathbf{W}_v \in \mathbb{R}^{d \times d}, \mathbf{b}_q, \mathbf{b}_k, \mathbf{b}_v \in \mathbb{R}^{d}$ are the weights and biases for querys, keys and values to obtain representations $\mathbf{Q}, \mathbf{K}, \mathbf{V} \in \mathbb{R}^{n \times d}$, respectively.

For head entity tagging, we concatenate the sentence representation $\mathbf{H}$ and head entity auxiliary features $\mathbf{H_h}$ as $[[\mathbf{w}_1, \mathbf{w}_1^h], ..., [\mathbf{w}_n, \mathbf{w}_n^h]]$. Then we calculate the score of the $i$-th token as the start and end position of a head entity with type $t_j \in \mathcal{T}$, respectively:
\begin{eqnarray}
     o_{ij}^{h_s} &=& \mathbf{W}_{ij}^{h_s} \cdot [\mathbf{w}_i, \mathbf{w}_i^h] + \mathbf{b}_{ij}^{h_s} \\
     o_{ij}^{h_e} &=& \mathbf{W}_{ij}^{h_e} \cdot [\mathbf{w}_i, \mathbf{w}_i^h] + \mathbf{b}_{ij}^{h_e}
\end{eqnarray}
where $o_{ij}^{h_s}$ represents the score that the $i$-th token is the start of the head entity with type $t_j$. 

Intuitively, different positions should have different thresholds, e.g., the threshold at entity boundary could be different from the boundary of other words.
And tuning a global threshold leads to suboptimal results because of the dependent on validation set data distribution, which leads to unsatisfactory test results and poor generalization. 
To improve the tagging accuracy and model generalization, we use the adaptive thresholding strategy to learn thresholds automatically. 
The tagger sets a position to 1 only when it has a higher score than its position-dependent threshold. 
For all start positions $p_{ij}^s$ of head entities with their types $t_j \in \mathcal{T}$ at position $i \in [1, n]$, we denote the positive classes (i.e. the positions expected to be tag to 1) as $\mathcal{P}_i$, negative classes as $\mathcal{N}_i$. And we set a ${AT}$ class to store all the start positions with respect to ${AT}$ type, where the position $i$ is denoted as $p_{i;{AT}}$. During training, the start tagging loss is as follows:
\begin{equation}
\begin{split}
    \mathcal{L}_h^s = &- \sum_{i=1}^{n} \sum_{{p_{ij}^s} \in \mathcal{P}_i} \log (\frac{\exp{(o_{ij})}}{\sum_{p_{ik}^s \in \mathcal{P}_i \cup \{p_{i;{AT}}\}} \exp{(o_{ik})}}) \\
    & - \sum_{i=1}^{n} \, \, \log (\frac{\exp{(o_{i;{AT}})}}{\sum_{p_{ik}^s \in \mathcal{N}_i \cup \{p_{i;{AT}}\}} \exp{(o_{ik})}})
\end{split}
\end{equation}
where the $o_{i;{AT}}$ represents the threshold at position $i$. Note that the start and end tagging requires different thresholds and follow the same loss function. The head entity tagging loss is the sum of start tagging loss $\mathcal{L}_h^s$ and end tagging loss $\mathcal{L}_h^e$:
\begin{equation}
    \mathcal{L}_h = \mathcal{L}_h^s + \mathcal{L}_h^e
    \label{lossfunction}
\end{equation}

Finally, we need to identify the relations and tail entities. 
We set an entity type embedding layer $\mathbf{V}_t \in \mathbb{R}^{K \times d_t}$ and a relative position embedding layer $\mathbf{V}_p \in \mathbb{R}^{n \times d}$. 
First, we obtain the head entity start and end token features $\mathbf{w}_a$, $\mathbf{w}_b$ from $\mathbf{H}$, derive the head entity type feature $\mathbf{w}^t$ from $\mathbf{V}_t$ and the relative position features $\mathbf{w}_a^p$, $\mathbf{w}_b^p$ from $\mathbf{V}_p$. Then the $(\mathbf{w}_a + \mathbf{w}_a^p)$, $(\mathbf{w}_b + \mathbf{w}_b^p)$ and $\mathbf{w}^t$ are concatenated to form the head entity features $\mathbf{w}^h$. 
Through the type-relation mapping mechanism, we determine the potential relation set $\mathcal{R}^{'}$ based on current entity type, aiming to help identify the tail entity more accurately by incorporating entity type information.
We concatenate the sentence representation $\mathbf{H}$, tail entity auxiliary features $\mathbf{H}_t$ and head entity features $\mathbf{w}^h$ as $[[\mathbf{w}_1, \mathbf{w}_1^t, \mathbf{w}^h], ..., [\mathbf{w}_n, \mathbf{w}_n^t, \mathbf{w}^h]]$. 
Consequently, we calculate the score of the $i$-th token as the start and end position of a tail entity with potential relation $r_j \in \mathcal{R}^{'} \subset \mathcal{R}$, respectively:
\begin{eqnarray}
     o_{ij}^{t_s} &=& \mathbf{W}_{ij}^{t_s} \cdot [\mathbf{w}_i, \mathbf{w}_i^t, \mathbf{w}^h] + \mathbf{b}_{ij}^{t_s} \\
     o_{ij}^{t_e} &=& \mathbf{W}_{ij}^{t_e} \cdot [\mathbf{w}_i, \mathbf{w}_i^t, \mathbf{w}^h] + \mathbf{b}_{ij}^{t_e}
\end{eqnarray}
where $o_{ij}^{t_s}$ represents the score that the $i$-th token is the start of the tail entity with relation $r_j$. Similar to head entity tagging loss, we directly calculate the tail entity tagging loss $\mathcal{L}_t$ according to Equation (\ref{lossfunction}).

Let $\mathcal{D}$ denote all the sentences, and $\mathcal Z_i$ denote all the relational triples in sentence $\mathcal{S}_i$. 
The loss function $\mathcal{L}$ is composed of two parts as follow:
\begin{equation}
    \mathcal{L} = \frac{1}{|\mathcal{D}|} \, (\sum_{\mathcal{S}_i \in \mathcal{D}} \sum_{h_j \in \mathcal Z_i} \mathcal{L}_{h_j} + \sum_{\mathcal{S}_i \in \mathcal{D}} \sum_{t_j \in \mathcal Z_i} \mathcal{L}_{t_j | h})
\end{equation}


\section{Experiments}
\subsection{Datasets, Baselines, and Settings}
We evaluate FastRE on three widely used public relation extraction datasets: NYT10~\cite{takanobu2019hierarchical}, NYT11~\cite{takanobu2019hierarchical} and NYT24~\cite{zeng2018extracting}. 
The statistics of these datasets are shown in Table~\ref{sd}.
We use [PER], [LOC], [ORG] and [OTH] to indicate the four entity types and  determine the mapping mechanism between entity types and predefined relations based on the semantics of each relation. 

We compare FastRE with the strong state-of-the-art relation extraction models:
(1) Tagging based models have been widely explored, such as NovelTagging~\cite{zheng2017joint}, CasRel~\cite{wei2020novel}, TPLinker~\cite{wang2020tplinker} and PRGC~\cite{zheng-etal-2021-prgc}.
(2) Generating based models convert the relation extraction task into a generation task, such as CopyRE~\cite{zeng2018extracting}, WDec~\cite{nayak2020effective}, and SPN~\cite{sui2021joint}.
(3) Other strong baselines cast the relation extraction task in a reinforcement learning perspective such as HRL~\cite{takanobu2019hierarchical}, or table filling framework like GRTE~\cite{ren-etal-2021-novel}.

The standard micro precision, recall, and F1 score are used to evaluate the results. To fairly compare with existing models, we follow previous work~\cite{takanobu2019hierarchical,wei2020novel} and use \emph{Partial Match} on all three datasets.

To evaluate the robustness and stability of the models, we use the same hyper-parameter setting on all datasets. 
Specifically, FastRE uses the 300-demensional GloVe embeddings~\cite{pennington2014glove} as the initial word vectors with $d$=128 hidden dimension. The kernel size $k_s$ and type embedding dimension $d_t$ is 3 and 64, respectively. 
FastRE applies 6 stacked $\rm{Block}(\cdot)$ and the dilation rate in each layer are 1, 2, 4, 1, 1, 1. Our model is optimized with AdamW~\cite{loshchilov2018decoupled} using learning rate 1e-3 with a linear warm up~\cite{goyal2017accurate} for the first 6\% steps followed by a linear decay to 0. We apply dropout~\cite{srivastava2014dropout} between convolutional layers with a rate 0.1. 
All the hyper-parameters reported in this work are based on the results on the validation sets. 
Other parameters are randomly initialized. Following CasRel~\cite{wei2020novel}, TPLinker~\cite{wang2020tplinker} and GRTE~\cite{ren-etal-2021-novel}, the max length of input sentences is set to 100. We train FastRE for 60 epochs with batch size 32, and all experiments are conducted on a NVIDIA RTX 2080Ti GPU.

\begin{table}[t]
\centering\setlength{\tabcolsep}{1mm}
\small 
\begin{tabular}{lcccc}
\toprule
\textbf{Dataset} & \textbf{\#Relation} & \textbf{Train} & \textbf{Valid} & \textbf{Test} \\
\midrule
{NYT10} & {29} & {70,339} & {-} & {4006}\\
{NYT11} & {12} & {62,648} & {-} & {369}\\
{NYT24} & {24} & {56,196} & {5,000} & {5,000}\\
\bottomrule
\end{tabular}
\caption{Statistics of datasets.}
\label{sd}
\end{table}


\begin{table*}[t]
\centering\setlength{\tabcolsep}{1mm}
\small 
\begin{tabular}{lccccccccccccccc}
\toprule
\multirow{2}{*}{\textbf{Models}} & \multicolumn{5}{c}{\textbf{NYT10}} & \multicolumn{5}{c}{\textbf{NYT11}} & \multicolumn{5}{c}{\textbf{NYT24}}\\
\cmidrule{2-16}
& {Prec.} & {Rec.} & {F1} & {Train.} & {Infer.} & {Prec.} & {Rec.} & {F1} & {Train.} & {Infer.} & {Prec.} & {Rec.} & {F1} & {Train.} & {Infer.}\\
\midrule
{NovelTagging~\cite{zheng2017joint}} & {59.3} & {38.1} & {46.4} & {-} & {-} & {46.9} & {48.9} & {47.9} & {-} & {-} & {62.4} & {31.7} & {42.0} & {-} & {-}\\
{CasRel~\cite{wei2020novel}$\dagger$} & {78.0} & {69.0} & {73.2} & {521} & {266} & {50.3} & {58.1} & {53.9} & {448} & {24} & {89.9} & {89.1} & {89.5} & {420} & {327}\\
{TPLinker~\cite{wang2020tplinker}$\dagger$} & {80.1} & {66.4} & {72.6} & {984} & {191} & {\textbf{56.2}} & {55.1} & {55.7} & {973} & {16} & {91.0} & {91.8} & {91.4} & {885} & {235}\\
{PRGC~\cite{zheng-etal-2021-prgc}$\dagger$} & {80.2} & {66.5} & {72.7} & {290} & {134} & {54.4} & {56.3} & {55.3} & {267} & {12} & {89.9} & {90.9} & {90.4} & {272} & {161}\\
\midrule
{CopyRE~\cite{zeng2018extracting}} & {56.9} & {45.2} & {50.4} & {-} & {-} & {34.7} & {53.4} & {42.1} & {-} & {-} & {61.0} & {56.6} & {58.7} & {-} & {-}\\
{WDec~\cite{nayak2020effective}} & \textbf{84.6} & {62.1} & {71.6} & {-} & {-} & {-} & {-} & {-} & {-} & {-} & \textbf{94.5} & {76.2} & {84.4} & {-} & {-}\\
{SPN~\cite{sui2021joint}$\dagger$} & {79.5} & {67.1} & {72.8} & {516} & {202} & {52.7} & {55.4} & {54.0} & {380} & {19} & {93.3} & {91.8} & {92.5} & {473} & {254}\\
\midrule
{HRL~\cite{takanobu2019hierarchical}} & {71.4} & {58.6} & {64.4} & {-} & {-} & {53.8} & {53.8} & {53.8} & {-} & {-} & {-} & {-} & {-} & {-} & {-}\\
{GRTE~\cite{ren-etal-2021-novel}$\dagger$} & {79.8} & {67.6} & {73.2} & {890} & {176} & {53.6} & {58.2} & {55.8} & {843} & {17} & {92.5} & \textbf{92.7} & \textbf{92.6} & {795} & {221}\\
\midrule
{FastRE (Ours)} & {78.0} & \textbf{70.1} & \textbf{73.8} & \textbf{100} & \textbf{18.4} & {54.1} & \textbf{58.7} & \textbf{56.3} & \textbf{87} & \textbf{1.6} & {89.6} & {86.3} & {87.9} & \textbf{96} & \textbf{23.0}\\
\bottomrule
\end{tabular}
\caption{Main Results. 
Baselines with $\dagger$ are produced with the source code provided in the original papers. The rest results of the baselines are retrieved from the original papers.  Results of all the models are the average of random three times. Train. and Infer. represent the total training (min) and inference time (s) on each dataset, respectively.
}
\label{mr}
\end{table*}

\begin{table*}[t]
\centering\setlength{\tabcolsep}{0.6mm}
\small 
\begin{tabular}{lccccccc}
\toprule
\multirow{2}{*}{\textbf{Models}} & \multirow{2}{*}{\textbf{Complexity}} & \multicolumn{2}{c}{\textbf{NYT10}} & \multicolumn{2}{c}{\textbf{NYT11}} & \multicolumn{2}{c}{\textbf{NYT24}}\\
\cmidrule{3-8}
& & {Param.} & {Infer. (1 / 8)} & {Param.} & {Infer. (1 / 8)} & {Param.} & {Infer. (1 / 8)}\\
\midrule
{CasRel~\cite{wei2020novel}} & $O(kn)$ & {107,729 K} & {66.2 / -} & {107,698 K} & {62.9 / -} & {107,720 K} & {65.3 / -}\\
{TPLinker~\cite{wang2020tplinker}} & $O(kn^2)$ & {109,626 K} & {104.7 / 47.5} & {109,548 K} & {97.5 / 42.5} & {109,603 K} & {103.2 / 47.0}\\
{SPN~\cite{sui2021joint}} & $O(cn)$ & {141,754 K} & {197.4 / 50.2} & {140,648 K} & {199.2 / 51.6} & {141,429 K} & {198.9 / 50.7}\\
{PRGC~\cite{zheng-etal-2021-prgc}} & $O(n^2)$ & {108,931 K} & {132.6 / 33.3} & {108,891 K} & {127.1 / 30.6} & {108,919 K} & {130.7 / 32.1}\\
{GRTE~\cite{ren-etal-2021-novel}} & $O(kn^2)$ & {119,450 K} & {97.5 / 43.8} & {119,166 K} & {98.8 / 44.2} & {119,387 K} & {98.3 / 44.1}\\
\midrule
{FastRE (Ours)} & $ O(mn) \to O(mn+kn)$ & {1,096 K} & {12.9 / 5.9 / \textbf{4.6}} & {1,092 K} & {11.7 / 5.5 / \textbf{4.4}} & {1,095 K} & {12.8 / 5.8 / \textbf{4.6}}\\
\bottomrule
\end{tabular}
\caption{Comparison of inference efficiency. Complexity are the theoretical decoding complexity with respect to sequence length $n$ and relation set size $k$. In SPN, $c$ is a constant independent of $k$ and typically less than $k$. In FastRE, $m$ denotes the number of entity types. Param. denotes the number of model parameters  obtained by the official implementation with default configuration. Infer. (1 / 8) denotes the inference time (ms) per instance with the batch size of 1 and 8. Results marked with \textbf{bold} denote the inference batch size is set to 128.}
\label{ce}
\end{table*}

\subsection{Main Results}
According to the main results shown in Table~\ref{mr}, FastRE is very effective compared with the state-of-the-art models based on BiLSTM and BERT, and achieves excellent performance on NYT10 and NYT11. 
Especially, FastRE outperforms the existing models in terms of training and inference speed on all datasets, which demonstrates the advantages of our convolutional encoder in efficiency. 
For example, the inference speed of FastRE is more than 7 $\times$ as PRGC, 15 $\times$ as CasRel and more than 10 $\times$ as other BERT based models. And the training speed of FastRE is 3 $\times$ as PRGC and more than 10 $\times$ as TPLinker. Note that except PRGC, the training time of other BERT based models varies greatly on different datasets. The training time required by FastRE on all datasets is close, although these datasets have different relation numbers. This is mainly because of introducing the mapping mechanism. From the results in PRGC which predicts relations first and extracts entity later, alleviating relation redundancy typically leads a relatively stable training time.
In addition, FastRE achieves significantly better performance on NYT10 and NYT11 compared to CasRel, indicating the effectiveness of the improved cascade binary tagging framework. 

We observe that the performance on NYT24 of FastRE is not good as the models based on the BERT. 
We believe that there exists two potential reasons for this phenomenon: (1) insufficient expression ability of FastRE, and (2) irrationality of the NYT24. 
First, FastRE employs a simple and efficient structure to approximate large Transformer based structures, so as to accelerate model training and inference speed, which limits its expression ability. 
Nonetheless, FastRE still outperforms BiLSTM based models such as WDec and closes to CasRel, means that the convolution based structure achieves comparable expression ability with pre-trained model BERT on RE task. 
In addition, the irrationality of the dataset would greatly impact the model performance. 
As~\cite{zhang-etal-2020-minimize} pointed out, over 90\% triplets in the NYT24 test set reoccurred in its training set, which easily overfitted by existing models and significantly limits the ability to verify the model generalization. 
We believe that these BERT based models gain a high score in NYT24 via memorizing the frequently reoccurred training set triplets because of the strong expression and fitting ability of BERT, which causes the overfitting instead of better generalization.

\subsection{Inference Efficiency}
According to the results shown in Table~\ref{ce}, the total parameters in FastRE are almost $1/100$ of other baselines, that ensures the faster forward calculation speed and back propagation update speed. Consequently, very few parameters can help FastRE achieve fast inference. 
Moreover, FastRE has lower theoretical decoding complexity compared to most models. 
When there are fewer entity types ($m < k$) and fewer average number of corresponding relations, FastRE has lower decoding complexity than CasRel. 
Although SPN has the lowest decoding complexity, the BERT based encoder and the non-autoregressive decoder become the most consuming parts and limit its inference efficiency. 

We set the batch size of all BERT based models to 8 for evaluating parallel processing ability. Consistent with the results obtained by~\cite{ren-etal-2021-novel}, TPLinker, SPN and GRTE have similar inference speed. 
FastRE is significantly faster than other models during inference. 
The results also indicate that the single-thread speed of FastRE is still much faster than other models. 
Similarly, although CasRel takes the longest time in whole inference stage, it is significantly faster than other BERT based models when processing a single instance, which shows the efficiency of cascade binary tagging framework. 
When the batch size of FastRE is 8, the speed of FastRE is about 5.5$\times$ as PRGC. 
And we can increase the batch size to 128 because of the simple model structure and the small parameter scale to obtain faster inference speed, which shows the superior parallel processing ability.

\subsection{Convergence Efficiency}
\begin{figure}[t]
	\centering
	\includegraphics[width=0.495\textwidth]{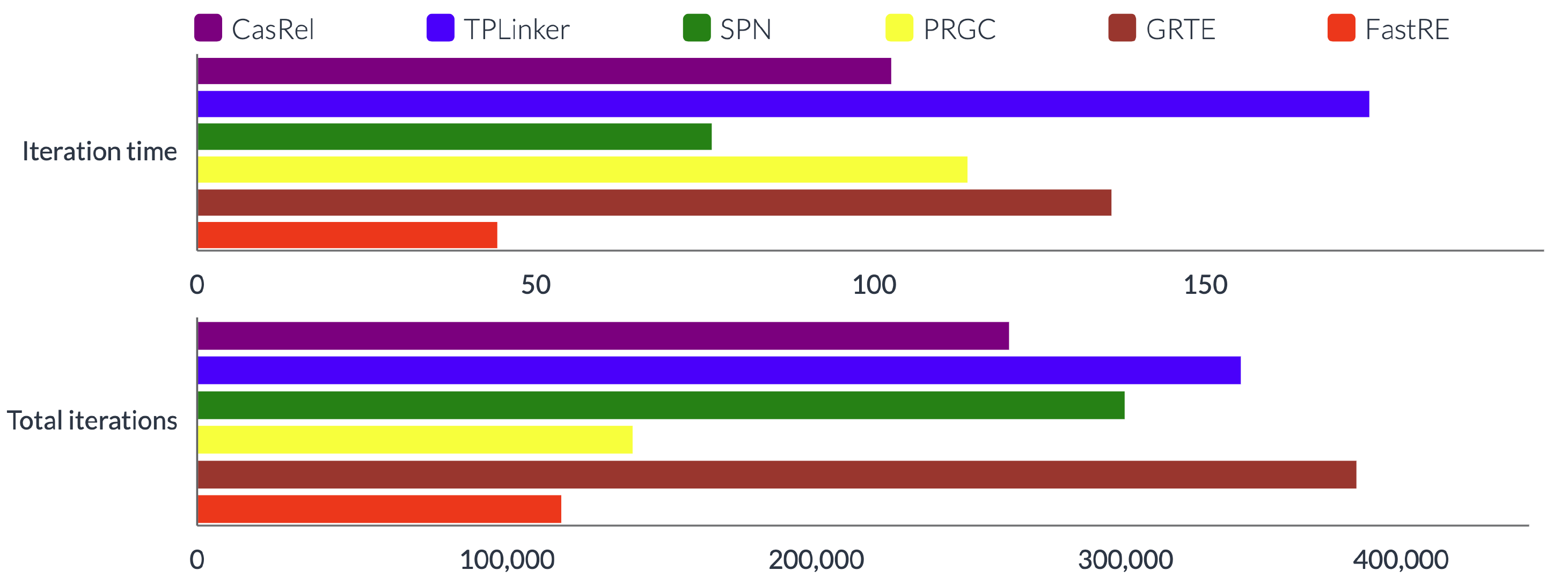}
	\caption{Iteration time (ms) and total iterations on NYT11. 
	}
	\label{time}
\end{figure}

To evaluate the model convergence efficiency, we analyze the iteration time and total iterations of different models on NYT11 as shown in Figure~\ref{time}. 
As different models require different batch sizes for training, we use the official implementation with default batch size configuration. 
Note that some models such as CasRel and TPLinker set the total epochs to 100 but could perform early stopping when the model performance no longer improving. 
For example, CasRel typically converges within 30 epochs. 
Therefore, we only count the time when a model converge, not the total number of epochs, which is quite different with another analysis~\cite{ren-etal-2021-novel}. 
We believe that our evaluation can more accurately reflect the real convergence efficiency.

The results show that FastRE outperforms other models in both aspects.
FastRE consumes the least iteration time because it requires the least computation with least parameters. Note that the iteration time in SPN is less than other BERT based models, but it requires much more iterations to converge as it casts the RE task to a complex bipartite matching problem and contains the most parameters compared to other models. Similar drawbacks occur in GRTE, which performs fussy table-filling with an iterative way and requires the most iterations among all the models.
Intuitively, FastRE should have few iterations for convergence with far less parameters than the BERT based model. 
However, the results show that FastRE has less advantage in the number of total iterations compared with PRGC.
This phenomenon motivates us to adopt more effective training strategies in the future.

\subsection{Structure Search}

We conduct structure search experiment as shown in Table~\ref{ss}. 
Setting the stacked block numbers to 3 greatly accelerates the model speed, but brings serious performance degradation. 
By changing the dilation rate of different blocks, we observe that the appropriate dilation rate plays an important role in improving the final performance. 
For example, when the dilated convolution is not used (the dilation rate is set to 1) or the dilation rate is set too large, the performance suffers grave decline. 
This is because dilated convolution is hard to capture long-distance dependencies by simply increasing dilation rate exponentially.
The few parameters of CNN fail to capture so much information and blindly increasing the receptive region will lead to worse results. 
Therefore, after a certain number of dilated convolutional layers, we replace the rest layers with vanilla convolutional layers to narrow the receptive region for fine tuning. 
In addition, continue increasing the number of stacked blocks only brings little performance improvement, but greatly slows down the model convergence speed and inference speed.
Therefore, FastRE adopts 6 stacked blocks and sets the dilation rate to 1, 2, 4, 1, 1, 1,  which is a structure that balances speed and performance. 

\begin{table}[t]
\centering\setlength{\tabcolsep}{0.5mm}
\small 
\begin{tabular}{lcccc}
\toprule
\textbf{Models} & \textbf{Param.} & \textbf{Train.} & \textbf{Infer.} & \textbf{F1} \\
\midrule
{FastRE ($L$=6, $d_i$=1, 2, 4, 1, 1, 1) } & {1096 K} & {100} & {18.4} & {73.8}\\
{$L$=3, $d_i$=1, 2, 4} & {800 K} & {50} & {13.7} & {68.6}\\
{$L$=6, $d_i$=1, 1, 1, 1, 1, 1} & {1096 K} & {100}  & {17.5} & {71.5}\\
{$L$=6, $d_i$=1, 2, 4, 8, 16, 32} & {1096 K} & {100} & {18.9} & {70.9}\\
{$L$=9, $d_i$=1, 2, 4, 1, 2, 4, 1, 1, 1} & {1392 K} & {150} & {28.3} & {74.1}\\
\bottomrule
\end{tabular}
\caption{Structure search on NYT10. Param. denotes the number of model parameters. Train. denotes the total training time (min) until convergence, and Infer. denotes the total inference time (s). For example, $L$=3, $d_i$=1, 2, 4 represents the encoder has 3 stacked blocks and the dilation rate is 1, 2, 4, respectively.}
\label{ss}
\end{table}

\begin{table}[t]
\centering\setlength{\tabcolsep}{1mm}
\small 
\begin{tabular}{lcccc}
\toprule
\textbf{Models} & \textbf{NYT10} & \textbf{NYT11} & \textbf{NYT24}\\
\midrule
{FastRE} & {73.8} & {56.3} & {87.9}\\
{- Dilated Convolution} &  {71.5} & {54.8} & {86.3}\\
{- Gated Unit} & {69.4} & {53.2} & {84.2} \\
{- Residual Connection} & {70.8} & {54.4} & {85.9} \\
{- Mapping Mechanism} & {72.2} & {54.4} & {86.4}\\
{- Adaptive Thresholding} & {73.0} & {55.5} & {87.0}\\
\bottomrule
\end{tabular}
\caption{Ablation study of components in FastRE.}
\label{as}
\end{table}

\subsection{Ablation Study}
We conduct ablation experiments to demonstrate the effectiveness of each component in FastRE with F1 scores (\%) reported in Table~\ref{as}. 
It is important to note that the impact of each component on efficiency is not obvious (except adaptive thresholding), so we only discuss the ablation results on performance.

\paragraph{Dilated Convolution} We set all dilation rates to 1. The dilated convolution can capture more local context information with less parameters and more speed. Without dilated convolutions, the ability to capture long-distance dependencies decreases significantly, resulting in a decline in the overall performance.

\paragraph{Gated Unit} The gated unit part ($\otimes {\rm{sigmoid}}(\mathbf{Y}_b)$) in Equation~(\ref{4}) is removed. The purpose of using gated unit is controlling information flow and selecting the most important information dynamically. Removing the gated units has a greater performance impact than removing dilated convolution and residual connection.

\paragraph{Residual Connection} We remove the residual connection part ($+ \mathbf{X}$) in Equation~(\ref{4}). Generally, the residual connection can prevent gradient vanishing, but it still has a positive impact on performance. The residual connection transmits information in multiple channels in multi-layer neural networks, resulting in simpler learning process and better generalization.


\paragraph{Mapping Mechanism} Without the mapping mechanism, namely, we set all entity types to [ANY], and [ANY] can map to all the relations, that causes a small drop in performance. However, one drawback of the mapping mechanism is that it can only deal with the case where the mapping from relation to entity type is uniquely determined. More general methods are worth exploring in the future.

\paragraph{Adaptive Thresholding} 
We perform sigmoid operation on all the positions and tuning a global threshold.
We find that using the global threshold improves the F1 scores on validation sets. However, the F1 scores on test sets with the global threshold are nearly 1\% lower than the results under the adaptive thresholding, which indicates that using adaptive thresholding makes the model have better generalization. We also observe that the convergence speed using adaptive thresholding loss is twice that using cross entropy loss. 
Sigmoid operation is simple but suffers from imbalanced classes issue as it produces plenty of 0 positions which affects the convergence speed. 
Rank-based adaptive thresholding loss balances the contribution of positive and negative classes in loss, making the training easier and faster.

\subsection{Discussion}
\paragraph{Other baselines and benchmarks}
Some latest work such as REBEL~\cite{huguet-cabot-navigli-2021-rebel-relation} and \textsc{ReRe}~\cite{xie-etal-2021-revisiting} are not addressed in baseline experiments.
Both models have comparable performance but lower efficiency compared to FastRE. 
REBEL (BART-large as backbone) cannot run in our experimental settings as it requires more computing resources. 
The parameters in REBEL are 400 $\times$ as FastRE and REBEL mainly focuses on performance rather than efficiency, which are different from the goal of our work. 
FastRE achieves the comparable performance as \textsc{ReRe} on NYT10 (73.8 vs 73.8) and NYT11 (56.3 vs 56.2), while the inference speed is 20 $\times$ faster than \textsc{ReRe}. 
The reason is that \textsc{ReRe} adopts pipeline extraction and uses two separate BERT models in relation classification and entity extraction. 
Furthermore, we have conducted systematic experiments on other benchmarks such as ADE and WebNLG, and the results consistently support the findings of our work. Due to space limitation, we cannot report all results.

\paragraph{Trade off between performance and efficiency}
The goal of FastRE is to accelerate RE efficiency while maintaining good performance. 
It employs a simple architecture to achieve a balance between performance and efficiency, because increasing the model complexity will decrease the efficiency. 
We examined the NYT24 and found that 90.49\% triplets overlapping in test set and training set, while it is only 72.64\% and 41.62\% in NYT10 and NYT11. This would leads BERT based models and Seq2seq based models to gain high scores because of rote memorization and exposure bias. Therefore, the results on NYT24 is not suitable to test the model generalization. 
Finally, we directly replaced the BERT encoder in CasRel with our convolutional encoder and the results show that the performance drops 1.8\%, 0.7\%, 3.3\% on three benchmarks, which shows that the representation ability of convolutional encoder is not as good as BERT on RE task. 
To sum up, FastRE has somewhat limited representation ability as its simplicity, but it achieves better generalization and is more suitable and applicable for unbiased datasets like NYT11 rather than NYT24.

\paragraph{Explainability}
FastRE improves training and inference speed while keeps promising performance, and it could be somewhat explained by the gradient techniques. Firstly, we visualized the distribution of attention scores on two separate self-attention layers, and found that the score of entity boundary is significantly higher than that of other words in both layers, which highlights the boundary features for subsequent entity prediction. 
Secondly, we collected the gradient of the loss w.r.t the input and computed the norm of the gradient for each word vector in the input. 
After normalization on each word gradient norm, we found that the important predicates in sentences (which indicate the relationships between entities) always receive higher (relative) gradient norms, while trivial conjunctions and prepositions have gradient norms close to 0.

\section{Conclusion}
This paper proposes a simple and fast relation extraction model, FastRE, to significantly reduce training and inference time but retain competitive performance. 
We propose a novel convolutional encoder combined with dilated convolution, gated unit and residual connection for RE.
We also improve the cascade binary tagging framework with type-relation mapping mechanism and position-dependent adaptive thresholding.
Experimental results show that FastRE achieves faster training and inference speed compared to the state-of-the-art models, while the performance is competitive.

\section*{Acknowledgements}
We thank the anonymous reviewers for their insightful comments.
The work was supported in part by the 13th Five-Year All-Army Common Information System Equipment Pre-Research Project (Grant Nos. 31514020501, 31514020503). 
All opinions are of the authors and do not reflect the view of sponsors.

\bibliographystyle{named}
\bibliography{FREE}

\end{document}